\documentclass{bmvc2k}

\usepackage{bm}
\usepackage{amssymb}
\usepackage{multirow}
\usepackage{booktabs}
\usepackage{colortbl}
\usepackage{arydshln}
\usepackage{hyperref}
\newcommand{\tablestyle}[2]{\setlength{\tabcolsep}{#1}\renewcommand{\arraystretch}{#2}\centering\footnotesize}

\title{In the Eye of Transformer: Global-Local Correlation for Egocentric Gaze Estimation}

\addauthor{Bolin Lai}{bolin.lai@gatech.edu}{}
\addauthor{Miao Liu*}{mliu328@gatech.edu}{}
\addauthor{Fiona Ryan}{fkryan@gatech.edu}{}
\addauthor{James M. Rehg*}{rehg@gatech.edu}{}

\addinstitution{
 College of Computing\\
 Georgia Institute of Technology\\
 Atlanta, GA
}

\runninghead{Bolin Lai et al.}{Transformer for Egocentric Gaze Estimation}

\begin{document}

\maketitle
\let\thefootnote\relax\footnotetext{* Equal corresponding author.}

\begin{abstract}
In this paper, we present the first transformer-based model to address the challenging problem of egocentric gaze estimation. We observe that the connection between the global scene context and local visual information is vital for localizing the gaze fixation from egocentric video frames. To this end, we design the transformer encoder to embed the global context as one additional visual token and further propose a novel Global-Local Correlation (GLC) module to explicitly model the correlation of the global token and each local token. We validate our model on two egocentric video datasets -- EGTEA Gaze+ and Ego4D. Our detailed ablation studies demonstrate the benefits of our method. In addition, our approach exceeds previous state-of-the-arts by a large margin. We also provide additional visualizations to support our claim that global-local correlation serves a key representation for predicting gaze fixation from egocentric videos. More details can be found in our website (\url{https://bolinlai.github.io/GLC-EgoGazeEst}).
\end{abstract}

\section{Introduction}
Findings in cognitive neuroscience suggest that eye movements reflect cognitive processes~\cite{yarbus2013eye}, which are essential for understanding human intention during daily activities~\cite{hayhoe2005eye}. Such an understanding of visual attention and intention can be valuable for many applications, including Augmented Reality (AR), Virtual Reality (VR), and Human-Robot Interaction (HRI). While wearable eye trackers are a standard way to obtain measurements of gaze behavior, they require calibration, consume significant power, and add substantial cost and complexity to wearable platforms. Alternatively, prior works~\cite{li2013learning, li2021eye, huang2018predicting, soo2015social, huang2020ego, tavakoli2019digging, thakur2021predicting, al2019ogaze, huang2020mutual, naas2020functional} seek to estimate the visual attention of the camera-wearer from videos captured from a first-person perspective. This task is known as egocentric gaze estimation.


\begin{figure}[tbp]
\centering
\includegraphics[width=1.0\linewidth]{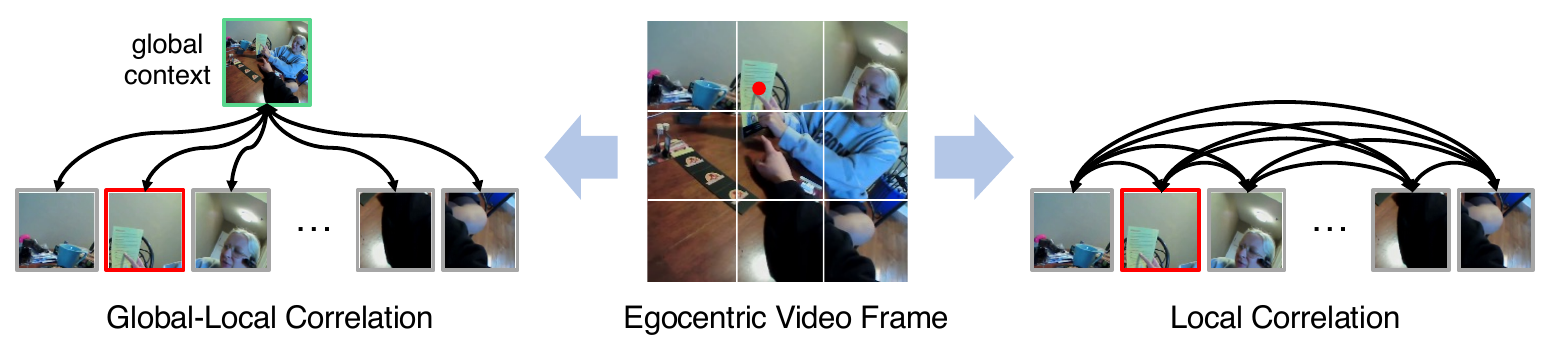}
\vspace{-1.5em}
\caption{Example of local correlation and global-local correlation for the task of egocentric gaze estimation (predicting where the camera-wearer is looking using egocentric video alone). The red dot represents the gaze ground truth (from a wearable eye tracker) and the image patch that contains the gaze target has red edges. Global-local correlation models the connections between the global context and each local patch, making it possible to capture, e.g., the camera wearer and social partner are pointing at the salient object. In contrast, local-local correlations may not yield an effective representation of the scene context.}
\vspace{-1.5em}
\label{fig:teaser}
\end{figure}

The key challenge in egocentric gaze estimation is to effectively integrate multiple gaze cues into a \emph{holistic} analysis of visual attention. Cues include the likelihood that different scene objects are gaze targets (i.e. salience), the relative location of gaze targets within the video frame (i.e. center prior), and the patterns of camera movement that are reflective of visual attention (i.e. head motions accompanying a gaze shift). Recently, the transformer has achieved great success in various vision tasks by modeling the spatio-temporal correlation among local visual tokens~\cite{strudel2021segmenter, lou2021transalnet, liu2021visual, dai2021up, fang2021you, ren2022shunted, yang2021focal, lee2022mpvit}. However, the pairwise comparisons performed by standard transformer Self-Attention (SA) are not optimized for interpreting local video features in the context of the global scene. The example Fig.~\ref{fig:teaser} illustrates the key role of comparisons between local patches and global context - the gaze target is a salient object pointed at by both the camera wearer and another person. Such a salient object can not be easily localized by only modeling the correlation of local patches. 

This paper introduces a novel transformer-based deep model that explicitly embeds global context and calculates spatio-temporal global-local correlation for egocentric gaze estimation. Specifically, we design a transformer encoder that adopts a global visual token embedding strategy to incorporate the global scene context. We then introduce a novel Global-Local Correlation (GLC) module that highlights the connection between global and local visual tokens. Finally, we adopt a transformer-based decoder to produce gaze prediction output. We evaluate our approach on two egocentric video datasets -- EGTEA Gaze+~\cite{li2018eye} and Ego4D~\cite{grauman2022ego4d}. Our proposed model is easy to incorporate into existing transformer-based video analysis architectures, and we show that it yields an \emph{improvement of more than $3.9\%$ in F1 score} over SOTA methods for egocentric gaze estimation. The code and pretrained models will be made publicly available to the research community. In summary, this work makes the following contributions:
\vspace{-0.6em}
\begin{itemize}
\item We introduce the first transformer-based approach to address the challenging task of egocentric gaze estimation.
\vspace{-0.5em}
\item We utilize a global visual token embedding strategy to incorporate global visual context into self-attention, and further introduce a novel Global-Local Correlation module to explicitly model the correlation between global context and each local visual token.
\vspace{-1.6em}
\item Our novel design obtains consistent improvement on the EGTEA Gaze+~\cite{li2018eye} and Ego4D~\cite{grauman2022ego4d} datasets and outperforms previous state-of-the-art methods by at least 3.9\% on EGTEA and 5.6\% on Ego4D in F1 score. Importantly, this is the first work that uses the Ego4D dataset for egocentric gaze estimation, which serves as important benchmark for future research in this direction.
\end{itemize}

\section{Related Work}
The task of egocentric gaze estimation, is distinct from prior work on eye tracking~\cite{krafka2016eye, macinnes2018wearable, ye2012detecting} and gaze target prediction from the third person video \cite{chong2020detecting, chong2018connecting, gaze360_2019, Nonaka_2022_CVPR}. In the interest of space, we limit our discussion to prior work on egocentric gaze estimation and related works on transformer-based video representation learning and video saliency prediction.

\noindent\textbf{Egocentric Gaze Estimation}.
Previous works focuses on analyzing human daily activities from egocentric videos~\cite{li2013learning, li2021eye, liu2019attention, huang2018predicting, liu2020forecasting, soo2015social, huang2020ego, tavakoli2019digging, zhang2017deep, thakur2021predicting, al2019ogaze, huang2020mutual, naas2020functional, jia2022generative, liu2021egocentric, 9665898}. Here, we discuss the most relevant works that develop deep models for egocentric gaze estimation. Zhang et al.~\cite{zhang2017deep} used deep models and an adversarial network to predict egocentric gaze location in future video frames. Huang et al.~\cite{huang2018predicting} incorporated temporal attention transition into saliency-based models for gaze estimation. Tavakoli et al.~\cite{tavakoli2019digging} studied both top-down and bottom-up cues that contribute to gaze guidance. Park et al.~\cite{soo2015social} introduced the novel problem of predicting joint attention during social interaction using egocentric videos. Huang et al.~\cite{huang2020ego} collected a new egocentric video dataset and developed a graphical model to detect joint attention. Thakur et al.~\cite{thakur2021predicting} proposed a multi-modal network that uses both video and inertial measurement unit data for more accurate egocentric gaze estimation. Naas et al.~\cite{naas2020functional} developed a tiling scheme for gaze prediction which enables a more efficient VR content delivery. Importantly, we are the first to develop a transformer-based architecture to address the problem of egocentric gaze estimation.

\noindent\textbf{Vision Transformer}.
Recently, vision transformers~\cite{dosovitskiy2020image} have demonstrated superior performance on image classification ~\cite{dai2021coatnet, liu2021swin, wang2021pyramid, ren2022shunted, yang2021focal, lee2022mpvit}, detection~\cite{dai2021dynamic, carion2020end, dai2021up, fang2021you}, segmentation~\cite{strudel2021segmenter, wang2021max, zheng2021rethinking, cheng2022masked, zhang2022topformer}, saliency prediction~\cite{ma2022video, lou2021transalnet, liu2021visual} and video analysis~\cite{arnab2021vivit, neimark2021video, fan2021multiscale, li2022mvitv2, bertasius2021space, wang2022deformable}. In this section, we focus on reviewing previous works that use vision transformers for pixel-wise visual prediction and video understanding. Strudel et al.~\cite{strudel2021segmenter} developed the first transformer-based architecture for semantic segmentation. Cheng et al.~\cite{cheng2022masked} further unified semantic, instance, and panoptic segmentation in one transformer architecture. Bertasius et al.~\cite{bertasius2021space} proposed TimeSformer for video action recognition. A similar idea is also explored in~\cite{arnab2021vivit}. Fan et al.~\cite{fan2021multiscale} designed a multiscale video transformer balancing computational cost and action recognition performance. Ma et al.~\cite{ma2022video} expanded transformers to visual saliency forecasting by using self-attention to capture the correlation between past and future frames. Liu et al.~\cite{liu2021visual} built a transformer-based model to detect salient objects on RGB-D images. Inspired by these successful applications of transformer architectures, we present the first work that uses a vision transformer to address the challenging task of egocentric gaze estimation. In addition, we introduce the novel Global-Local Correlation (GLC) module that provides additional insight into video representation learning with self-attention.

\noindent\textbf{Visual Saliency}.
Visual saliency prediction has been well studied in computer vision~\cite{pan2017salgan, wang2017video, che2019gaze, wu2020salsac, kroner2020contextual, jia2020eml, sun2022visual, lou2021transalnet, wang2021spatio, chen2021video, khattar2021cross, bellitto2021hierarchical, tsiami2020stavis, jiang2022does}. We mainly review previous works on saliency prediction from videos. Wang et al.~\cite{wang2017video} expanded image saliency models to videos by incorporating a new branch to handle temporal information. Wu et al.~\cite{wu2020salsac} proposed SalSAC, which shuffles features of different CNN layers and feeds them to a correlation-based ConvLSTM. Wang et al.~\cite{wang2021spatio} used multiple spatio-temporal self-attention modules to address the limitation of fixed kernel size in 3D models and to model long-range temporal dependencies. Chen et al.~\cite{chen2021video} decomposed video saliency prediction into spatial pattern capture and spatio-temporal reasoning. Lou et al.~\cite{lou2021transalnet} combined a convolutional network and transformer architecture to model the long-range spatial context. While visual saliency prediction localizes interesting spatial regions as potential attention targets, egocentric gaze estimation seeks to determine the gaze target of the camera wearers as they interact with a scene. Moreover, the scene context captured from egocentric video is complex and rapidly changing, which requires a gaze estimation model with the ability of explicitly reasoning about the correlation between local visual features and global scene context. In our experiment section, we demonstrate that our proposed GLC module can significantly benefit gaze estimation performance under this challenging setting.

\begin{figure}[tbp]
\centering
\includegraphics[width=0.99\linewidth]{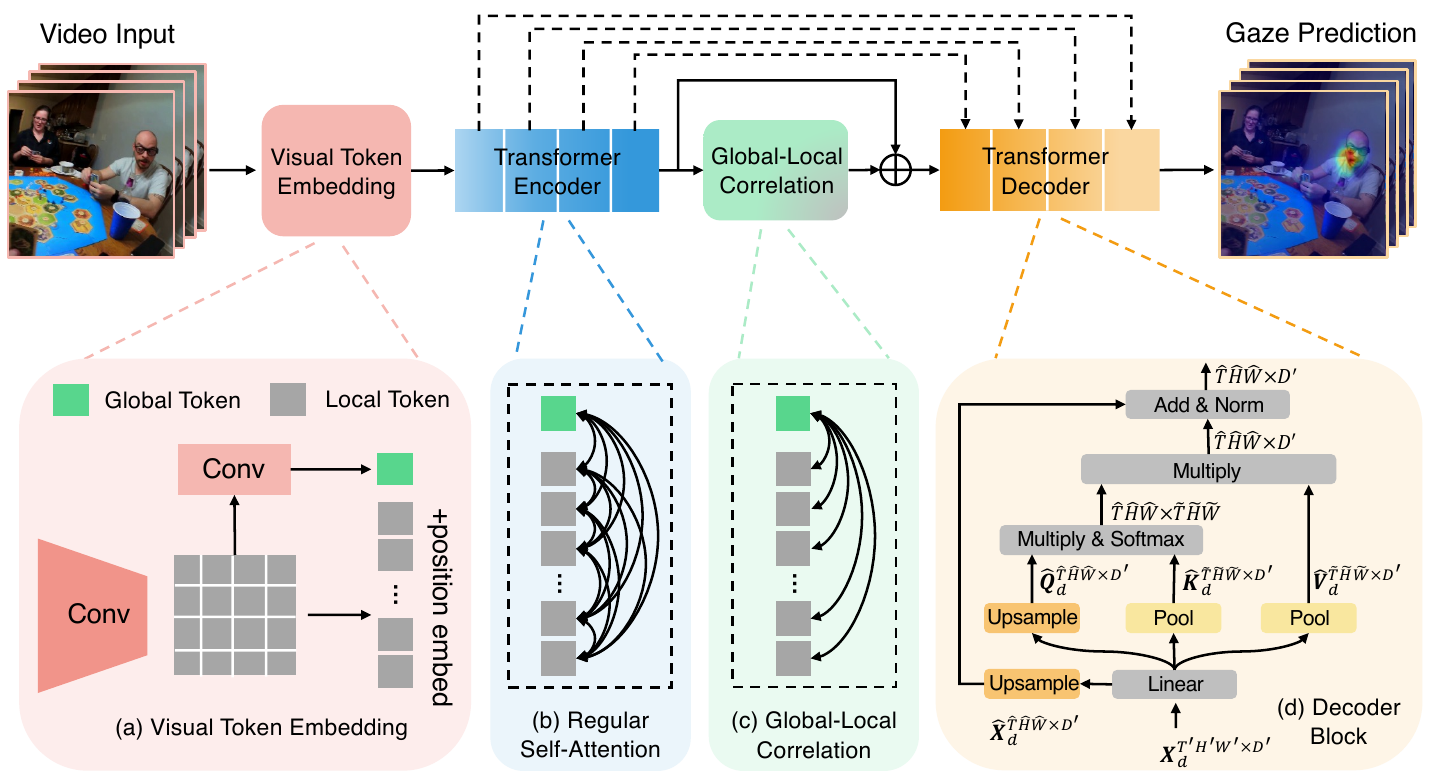}
\vspace{-1.0em}
\caption{Architecture of the proposed model. The model consists of four modules -- (a) \textbf{Visual Token Embedding Module} encodes the input into local tokens and one global token, (b) \textbf{Transformer Encoder} is composed of multiple regular self-attention and linear layers, (c) \textbf{Global-Local Correlation Module} models the correlation of global and local tokens, and (d) \textbf{Transformer Decoder} maps encoded video features from Transformer Encoder and GLC to gaze prediction. $\oplus$ denotes concatenation along the channel dimension.}
\vspace{-1em}
\label{fig:architecture}
\end{figure}

\section{Method}
\vspace{-0.3em}


Given an input egocentric video clip with fixed length $T$ and spatial dimension $H\times W$, our goal is to predict the gaze location in each video frame. Following~\cite{li2018eye}, we consider the gaze prediction as a probabilistic distribution defined on the $2D$ image plane.

Fig.~\ref{fig:architecture} presents an overview of our proposed method. We use the recent multi-scale video transformer (MViT) architecture~\cite{fan2021multiscale} as the backbone network for video representation learning. We extend MViT by designing the \emph{Visual Token Embedding Module} to generate the spatio-temporal tokens of both local visual patches and global visual context and feed them into the standard \emph{Multi-Head Self-Attention Module}. We then utilize a novel \emph{Global-Local Correlation (GLC) Module} to explicitly model the correlation between global and local visual tokens for gaze estimation. Finally, we make use of the \emph{Decoder Network} to predict the gaze distribution based on the learned video representation from the GLC module.




\subsection{Transformer Encoder with Global Visual Token Embedding}
\textbf{Visual Token Embedding}. We split the input video sequence into non-overlapping patches with size $s_T\times s_H\times s_W$ and adopt a linear mapping function to project each flattened patch into $D$-dimension vector space. Following MViT~\cite{fan2021multiscale}, this is equivalent to a convolutional layer with a stride of $s_T\times s_H\times s_W$ and a number of output channels of $D$. This operation results into $N$ tokens where $N=\frac{T}{s_T}\times\frac{H}{s_H}\times\frac{W}{s_W}$. In addition, the learnable positional embedding $\bm{E}\in\mathbb{R}^{N\times D}$ is added to the local tokens. Our key insight is to further embed global information into a global visual token using convolutional operations, as illustrated in Fig.~\ref{fig:architecture}(a). Since there is a single global token, it does not require positional embedding.
In our experiments, we
examine multiple strategies to embed the global visual token.




 \noindent\textbf{Multi-Head Self-Attention Module}. The $N$ local tokens and one global token are fed into a transformer encoder consisting of multiple self-attention blocks. The number of local tokens is downsampled after each self-attention block, while the number of global tokens remains $1$. Suppose the input of the $j$-th layer of encoder is $\bm{X}_e^{(j)}=[\bm{x}_i^{(j)}]_{i=1}^{N_j+1}\in\mathbb{R}^{(N_j+1)\times D_j}$, where $N_j$ is the number of local tokens, $D_j$ is the vector length of each token and $\bm{x}_i^{(j)}$ is the $i$-th row of $\bm{X}_e^{(j)}$ denoting the $i$-th token of size $1\times D_j$. For simplicity, we omit subscript and superscript of $j$ and multi-head operation in the following equations. In each self-attention layer, correlations are calculated in each token pair as shown in Fig.~\ref{fig:architecture}(b). They are used to reweight values of each token after softmax. Formally, we denote the query, key and value matrices of each self-attention layer in an encoder block as $\bm{Q}_e^{(N+1)\times D}=[\bm{q}_i]_{i=1}^{N+1}$, $\bm{K}_e^{(N+1)\times D}=[\bm{k}_i]_{i=1}^{N+1}$ and $\bm{V}_e^{(N+1)\times D}=[\bm{v}_i]_{i=1}^{N+1}$. The self-attention in transformer encoder is formulated as
\begin{equation}\label{eq:regular_sa}
    Attention(\bm{Q}_e, \bm{K}_e, \bm{V}_e) = Softmax(\bm{Q}_e\bm{K}_e^T / \sqrt{D})\bm{V}_e \in \mathbb{R}^{(N+1)\times D}.
\end{equation}
Finally, we attach a standard linear layer after the self-attention operation.





\subsection{Global-Local Correlation}
Even though global information has been explicitly embedded into the global visual token in our model, the transformer encoder treats the global and local tokens equivalently as shown in Eq.~\ref{eq:regular_sa} and Fig.~\ref{fig:architecture}(b). In this case, global-local correlation is diluted by correlations among the local tokens, limiting its impact on gaze estimation. In order to address this problem, we propose to increase the available capacity to model global-local token interactions. 
Our solution is a novel Global-Local Correlation module described in Fig.~\ref{fig:architecture}(c). 



Formally, we denote the global token as the first row vector of $\bm{X}_e$, $\emph{i.e.}, \bm{x}_1$. Thus $\bm{q}_1$, $\bm{k}_1$ and $\bm{v}_1$ are the query, key and value projected from the global token, respectively. To explicitly model the connection between global and local visual features, we only calculate the correlation between each local token and the global token, \emph{i.e.}, $Correlation(\bm{x}_i, \bm{x}_1)$, as well as its self-correlation, \emph{i.e.}, $Correlation(\bm{x}_i, \bm{x}_i)$. Then correlation scores are normalized by softmax to further re-weight the values. 
We exploit a suppression matrix~\cite{liu2021swin} $\bm{S}^{(N+1)\times (N+1)}$ to suppress the correlation of other tokens, where
\begin{equation}
    \bm{S}^{(N+1)\times(N+1)} = [s_{ij}],\quad s_{ij}=\left\{
    \begin{array}{ll}  
        0, &\mathrm{if}\; i=j \; \mathrm{or} \; j=1  \\  
        \lambda, &\mathrm{otherwise}.
    \end{array}
    \right.
\end{equation}
We assign zeros to the diagonal and the first column in $\bm{S}$ and set a large value $\lambda$ for the other elements. We follow the empirical choice from the implementation of~\cite{liu2021swin} and set $\lambda=10^8$ in our experiments. Formally, the proposed GLC can be formulated as the following:
\begin{equation}
    GLC(\bm{Q}_e, \bm{K}_e, \bm{V}_e) = Softmax((\bm{Q}_e\bm{K}_e^T - \bm{S}) / \sqrt{D})\bm{V}_e\in \mathbb{R}^{(N+1)\times D}
\end{equation}
In this way, we keep the values on the first column and the diagonal, and map them into probability distributions, while values in other positions are nearly ``masked out'' after the softmax. Residual connections and linear layers are also used in the GLC module as in the regular self-attention block. Finally, the output tokens from the GLC are concatenated with those from the transformer encoder in the channel dimension. We denote outputs of the GLC and the last encoder block as $\bm{X}_e^{GLC}\in\mathbb{R}^{(N+1)\times D}$ and $\bm{X}_e^{SA}\in\mathbb{R}^{(N+1)\times D}$. The concatenation can then be formulated as $\bm{X}_e = \bm{X}_e^{SA} \oplus \bm{X}_e^{GLC} \in \mathbb{R}^{(N+1)\times2D}$. The fused tokens $\bm{X}_e$ are subsequently fed into the transformer decoder for gaze estimation.

\subsection{Transformer Decoder}
To produce the gaze distribution with the desired spatio-temporal resolution, we adopt a decoder to upsample the encoded features. We utilize a transformer decoder based on the multiscale self-attention block of  MViT~\cite{fan2021multiscale}. Suppose each decoder layer takes visual features $\bm{X}_d\in\mathbb{R}^{T'H'W'\times D'}$ as inputs and the corresponding query, key and value matrices are $\bm{Q}_d^{T'H'W'\times D'}$, $\bm{K}_d^{T'H'W'\times D'}$ and $\bm{V}_d^{T'H'W'\times D'}$. As shown in Fig.~\ref{fig:architecture}(d), we replace the original pooling operation for the query matrix with an upsampling operation implemented with trilinear interpolation and keep the pooling for the key and value matrices. Following~\cite{fan2021multiscale}, $\widehat{\bm{Q}}_d$ is obtained by applying a deconvolutional operation on $\bm{Q}_d$, while $\widehat{\bm{K}}_d$ and $\widehat{\bm{V}}_d$ are obtained by applying convolutional operations on $\bm{K}_d$ and $\bm{V}_d$. Then, the output of self-attention is calculated in the same way as Eq.~\ref{eq:regular_sa}. In addition, we keep the skip connection in the self-attention layers and replace the pooling operation in skip connections with trilinear interpolation, which produces the upsampled output with dimension $\widehat{T}\widehat{H}\widehat{W}\times D'$. Our decoder is composed of 4 decoding blocks. Skip connections are used to combine intermediate features of the encoder with corresponding decoder features. Finally, another linear mapping function is used to output the final gaze prediction.



\subsection{Network Architecture and Model Training}
We adopt MViT~\cite{fan2021multiscale} as the backbone, with weights initialized from Kinetics-400 pretraining~\cite{kay2017kinetics}. The token embedding stride is set as $s_T=2$, $s_H=4$ and $s_W=4$ and the embedding dimension is $D=96$. The encoder is composed of 16 self-attention layers that are divided into 4 blocks. The number of tokens is downsampled at the transition between two blocks. We build the decoder with 4 decoder blocks corresponding to the 4 blocks in the encoder. After getting raw output from decoder, softmax is applied on each frame with a temperature $\tau$. This can be formally written as $\hat{p}_{ij} = \frac{\exp(\hat{y}_{ij} / \tau)}{\sum_{i,j}\exp({\hat{y}_{ij} / \tau)}}$
where $\hat{y}_{ij}$ is the logit at location $(i,j)$ from the model and $\hat{p}_{ij}$ is probability after softmax. In experiments, $\tau$ is empirically set as 2. We use KL-divergence loss to capture the difference between labels and predictions.  More details of the training parameters can be found in the supplementary.


\section{Experiment}
\subsection{Datasets and Metrics}
We conducted our experiments on two egocentric video datasets with gaze tracking data to serve as ground truth -- EGTEA Gaze+~\cite{li2018eye} and Ego4D~\cite{grauman2022ego4d}. The EGTEA dataset is captured under the meal preparation setting, which involves a great deal of hand-object interactions. We used the first train/test split from EGTEA in our experiments (8299 clips for training and 2022 clips for testing). The Ego4D dataset includes 27 videos of 80 participants totaling 31 hours with gaze tracking data captured under the social setting. We split the long videos into 5-second video clips and pick clips containing gaze fixation. More details of the selection of clips with gaze fixations are discussed in supplementary. We used 20 videos (15310 clips) for training and the other 7 videos (5202 clips) for testing. This is the first work that uses the Ego4D dataset for egocentric gaze estimation, and we will make our split publicly available to drive future research on this topic. Following~\cite{li2018eye, li2021eye}, we adopt F1 score, recall, and precision as the evaluation metrics.



\begin{table}[tbp]
\begin{center}
\begin{tabular}{lcccccc}
\toprule
\multirow{2}{*}{Methods}  & \multicolumn{3}{c}{EGTEA Gaze+} & \multicolumn{3}{c}{Ego4D} \\
\cmidrule(lr){2-4} \cmidrule(lr){5-7}
& F1 & Recall & Precision & F1 & Recall & Precision \\
\midrule
MViT              & 43.0 & 57.8 & 35.4 & 40.9 & 57.4 & 31.7 \\
MViT + (a)        & 43.4 & 58.4 & 34.5 & 41.5 & 56.8 & 32.6 \\
MViT + (b)        & 43.5 & 59.2 & 34.4 & 41.4 & 57.3 & 32.4 \\ 
MViT + (c)        & 43.7 & 58.3 & 34.9 & 41.3 & 57.5 & 32.2 \\
MViT + (d)        & 43.9 & 59.0 & 34.9 & 41.7 & 57.6 & 32.7 \\
\hdashline
MViT + (d) + SA   & 44.1 & 58.8 & \textbf{35.3} & 42.1 & \textbf{58.5} & 32.9 \\
\rowcolor[HTML]{FAEBD7} MViT + (d) + GLC  & \textbf{44.8} & \textbf{61.2} & \textbf{35.3} & \textbf{43.1} & 57.0 & \textbf{34.7} \\
\bottomrule
\end{tabular}
\end{center}
\vspace{-0.6em}
\caption{Evaluation of different global embedding approaches and global-local correlation module. (a)(b)(c)(d) are different global embedding strategies elaborated in Section~\ref{sec:results}. \emph{SA} and \emph{GLC} denote regular self-attention and global-local correlation module, respectively.}
\vspace{-1.8em}
\label{tab:ablation}
\end{table}

\subsection{Experimental Results}\label{sec:results}

\noindent\textbf{The Design Choice of Global Visual Embedding}.\
Our key insight is embedding the global visual information into the transformer architecture for egocentric gaze estimation. Here, we first explore 4 global visual embedding strategies -- (a) direct max pooling on the input, (b) max pooling on the unflattened local tokens, (c) convolutional layers on the input and (d) convolutional layers on the unflattened local tokens. 

As shown in Table~\ref{tab:ablation}, all four global embedding strategies improve the performance of vanilla MViT model on both the EGTEA dataset and the Ego4D dataset. This result supports our claim that global context is essential for gaze estimation. Among the four embedding strategies, (d) achieves the largest performance improvement on both datasets ($+0.9\%$ on EGTEA and $+0.8\%$ on Ego4D). This indicates that convolutional layers and the embedded local tokens can facilitate the learning of global context. Thus, we use this strategy in the following experiments. Note that all baseline methods use the same transformer decoder.

\noindent\textbf{Evaluation of Global-Local Correlation}. We also evaluate the Global-Local Correlation (GLC) module of our model. As presented in Table~\ref{tab:ablation}, our full model -- MViT+(d)+GLC outperforms the baseline MViT by $+1.8\%$ on EGTEA dataset and $+2.2\%$ on Ego4D dataset. Specifically, the GLC module contributes to a performance gain of $+0.9\%$ on EGTEA Gaze+ and $+1.4\%$ on Ego4D (comparing to MViT+(d)). This result suggests that the GLC can break down the mathematical equivalence of global and local tokens in regular self-attention, thereby ``highlighting'' the global-local connection in the learned representation.


\noindent\textbf{Does the Performance Improvement Come from Additional Parameters}?  It is possible that the performance of our model benefits from additional parameters in the GLC module. In Table~\ref{tab:ablation}, we report the results of another baseline model, where we replace the GLC module with a regular self-attention (SA) layer. Interestingly, the additional SA layer has minor influence on the overall performance ($+0.2\%$ on EGTEA and $0.4\%$ on Ego4D). In contrast, our model outperforms this baseline by $+0.7\%$ on EGTEA and $+1.0\%$ on Ego4D. This result indicates that the performance boost of our method does not simply come from the additional parameters of GLC. Instead, the explicit modeling of the connection between global and local visual features is the key factor in the performance gain.


\begin{table}[t]
\begin{center}
\begin{tabular}{lccc}
\toprule
Methods  & F1 & Recall & Precision \\
\midrule
Center Prior                                    & 10.7 & 32.0 & 6.4 \\
GBVS~\cite{harel2006graph}                      & 15.7 & 45.1 & 9.5 \\
EgoGaze~\cite{li2013learning}                   & 16.3 & 16.3 & 16.3 \\
SimpleGaze                                      & 31.3 & 41.8 & 16.1 \\ 
Deep Gaze~\cite{zhang2017deep}                  & 34.5 & 43.1 & 28.7 \\
Gaze MLE~\cite{li2021eye}                       & 26.6 & 35.7 & 21.3 \\
Joint Learning~\cite{li2021eye}                 & 34.0 & 42.7 & 28.3 \\
Attention Transition~\cite{huang2018predicting} & 37.2 & 51.9 & 29.0 \\
I3D-R50~\cite{feichtenhofer2019slowfast}   & 40.9 & 57.2 & 31.8 \\
\hdashline
MViT              & 43.0 & 57.8 & 35.4 \\
\rowcolor[HTML]{FAEBD7} Ours  & \textbf{44.8} & \textbf{61.2} & \textbf{35.3} \\
\bottomrule
\end{tabular}
\end{center}
\vspace{-0.8em}
\caption{Comparison with previous methods on EGTEA Gaze+. Our complete model is highlighted. The proposed model outperforms previous approaches by a significant margin.}
\vspace{-0.2em}
\label{tab:egtea}
\end{table}

\begin{table}[t]
\begin{center}
\begin{tabular}{lccc}
\toprule
Methods  & F1 & Recall & Precision \\
\midrule
Center Prior                                    & 14.9 & 21.9 & 11.3 \\
GBVS~\cite{harel2006graph}                      & 18.0 & 47.2 & 11.1 \\
Attention Transition~\cite{huang2018predicting} & 36.4 & 47.6 & 29.5 \\
I3D-R50~\cite{feichtenhofer2019slowfast}   & 37.5 & 52.5 & 29.2 \\
\hdashline
MViT               & 40.9 & 57.4 & 31.7
\\
\rowcolor[HTML]{FAEBD7} Ours  & \textbf{43.1} & \textbf{57.0} & \textbf{34.7} \\
\bottomrule
\end{tabular}
\end{center}
\vspace{-0.8em}
\caption{Comparison with previous methods on Ego4D. Our complete model is highlighted. The model shows consistent superiority over other state of the arts on all metrics.}
\vspace{-1.8em}
\label{tab:ego4d}
\end{table}

\noindent\textbf{Comparison with Previous State-of-the-Art}. In addition to these studies to evaluate the components of our model, we compare our approach with prior work. Results are presented in Table~\ref{tab:egtea} and Table~\ref{tab:ego4d}. Interestingly, the baseline MViT model easily outperforms all previous works that use CNN-based architectures on both the EGTEA dataset and the Ego4D dataset. In addition, our proposed method outperforms the best CNN model by $+3.9\%$ on F1, $+4.0\%$ on recall and $+3.5\%$ on precision. On Ego4D, our method surpasses the best CNN model by $+5.6\%$ on F1, $+4.5\%$ on recall and $+5.5\%$ on precision. These results demonstrate the superiority of using a transformer-based architecture for egocentric gaze estimation as well as the effectiveness and robustness of our proposed method. We note that the improvement of our model is more prominent on Ego4D than EGTEA Gaze+. We speculate that this is because the Ego4D videos with gaze tracking data are captured under social interaction scenarios that contain interactions with both people and objects, and thus require the model to more heavily consider the global-local connections (\emph{e.g.} the visual information about a social partner's gesture to an object) to predict the gaze. Another possible reason is that the Ego4D dataset has more samples to train the transformer-based model. 


\begin{figure}[t]
\centering
\includegraphics[width=0.9\linewidth]{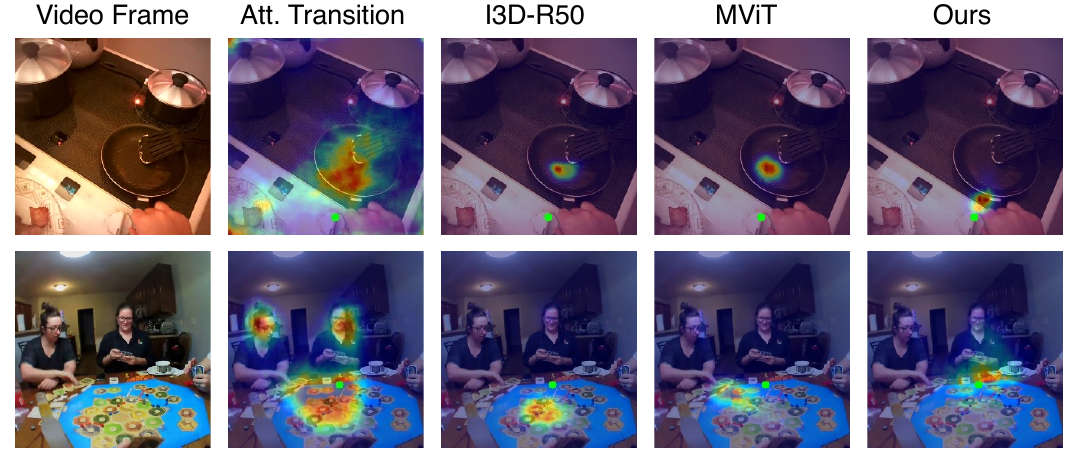}
\vspace{-1.2em}
\caption{Visualization of gaze estimation. The first sample is from EGTEA Gaze+ and the second is from Ego4D. Estimated gaze is represented as a heatmap overlayed on input frames. Green dots denote the ground truth gaze location.}
\vspace{-0.5em}
\label{fig:cmp_sota}
\end{figure}

\begin{figure}[t]
\centering
\includegraphics[width=\linewidth]{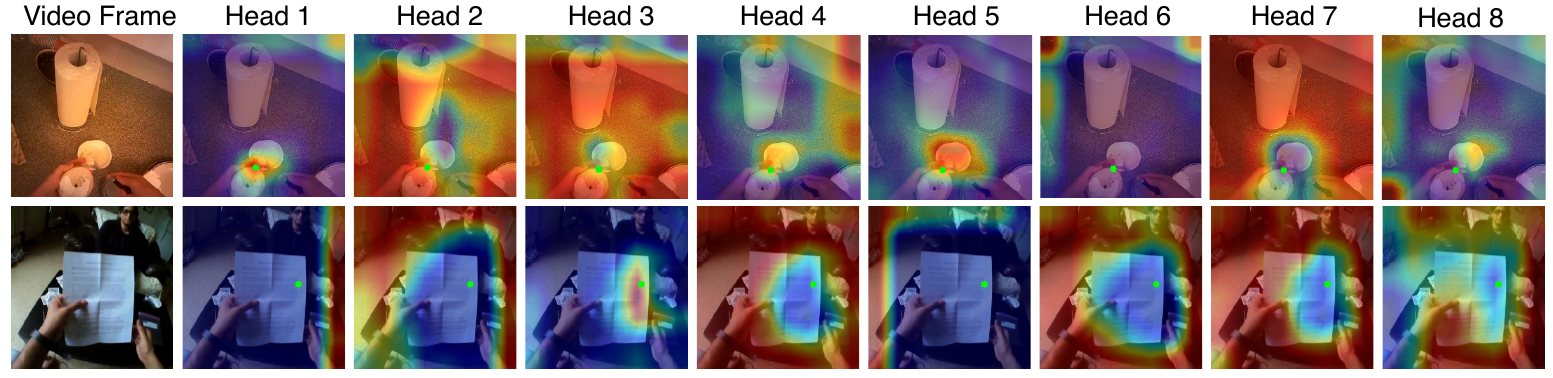}
\vspace{-1.8em}
\caption{Visualization of the eight heads in global-local correlation module. The first sample is from EGTEA Gaze+ and the second is from Ego4D. Green dots denote gaze location.}
\vspace{-1.3em}
\label{fig:glc_vis}
\end{figure}

\subsection{Remarks}
\noindent\textbf{Visualization of Predictions}.\
We visualize predictions of our model and other previous methods in Fig.~\ref{fig:cmp_sota}. Attention transition~\cite{huang2018predicting} tends to overestimate gaze area which includes more uncertainty and ambiguity. I3D-R50~\cite{feichtenhofer2019slowfast} and vanilla MViT~\cite{fan2021multiscale} architectures run into failure modes when there are multiple objects and people in the scene. In contrast, our model, by explicitly modeling the connection between the global and local visual tokens, more robustly predicts the egocentric gaze distribution from the input video clip.

\noindent\textbf{What has been learned by the Global-Local Correlation module}?\ We additionally empirically analyze our proposed GLC module. We first calculate the correlation of the global token and each local token, and then normalize the calculated weights into a probabilistic distribution. A higher score suggests that the GLC captures a stronger connection between the particular local token and the global context. We reshape and upsample these weight distributions to form a heatmap, which we overlay with the original input. Since the GLC module applies a multi-head operation, we visualize the results from different heads in Fig.~\ref{fig:glc_vis}. Interestingly, the correlations captured by the GLC heads are quite diverse. We observe that the GLC module does assign different weights to local tokens, thereby capturing the different global-local connections for each token. Another important finding is that some heads learn to attend to background pixels to prevent the model from omitting important scene context. We provide further commentary in the supplementary.
\vspace{-0.7em}


\section{Conclusion}

In this paper, we develop a transformer-based architecture to address the task of estimating the camera wear's gaze fixation based only on egocentric video frames. Our key insight is that our global visual token embedding strategy, which encodes global visual information into the self-attention mechanism, and our global-local correlation (GLC) module, which explicitly reasons about the connection between global and local visual tokens, facilitate strong representation learning for egocentric gaze estimation. Our experiments on the EGTEA Gaze+ and Ego4D datasets demonstrate the effectiveness of our approach. We believe our work serves as an essential step in analyzing gaze behavior from egocentric videos and provides valuable insight into learning video representations with transformer architectures.

\bibliography{bmvc_review}

\newpage
~\\
\centerline{\LARGE \textbf{Supplementary}}
~\\

\setcounter{section}{0}  

This is supplemental material for the paper titled "In the Eye of Transformer: Global-Local Correlation for Egocentric Gaze Estimation". We organize the content in the following way:

\begin{itemize}
\item A -- Data Processing
\item B -- Implementation Details
\item C -- Experiments on Action Recognition
\item D -- Details of Different Global Visual Embedding Strategies
\item E -- More Visualization Examples of Gaze Estimation
\item F -- Future Work

\end{itemize}

\renewcommand\thesection{\Alph {section}}
\renewcommand\thesubsection{\thesection.\arabic{subsection}}

\section{Data Processing}
\label{sec:data}
At training time, we randomly sample 8 frames from each video with a sampling interval of 8 as input (\emph{i.e.} selecting 8 frames from a 72-frame window with equal spacing). All videos are spatially downsampled to 256 in height while keeping the original aspect ratio. We further implement multiple data augmentations including random flipping, shifting, and resizing. We then randomly crop each frame to get an input with dimensions $8\times256\times256$. The output from the decoder is a downsampled heatmap with dimension $8\times56\times56$. For visualization, the output heatmap is  upsampled to match the input size by trilinear interpolation. At inference time, the input clip is center-cropped. For gaze labels, we generate a gaussian kernel centered at the gaze location in each input frame with a kernel size of 19 following~\cite{chong2020detecting}. We use a uniform distribution for frames where gaze is not tracked in training and only calculate metrics on frames with fixated gaze in testing as in~\cite{li2018eye}. For the EGTEA Gaze+~\cite{li2018eye} dataset, we determine which frames to calculate metrics on by using the provided label of gaze fixations and saccades. On the Ego4D~\cite{grauman2022ego4d} dataset, no label of gaze type is available. We calculate the euclidean spatial distance of gaze between adjacent frames and consider the tracked gaze to be a saccade if the distance is above a threshold, and treat it as fixation otherwise. We adopt an empirical threshold of 40.

\section{Implementation Details}
\label{sec:implementation}
We show the parameter details of each layer in Table~\ref{tab:architecture}. Data is input to the local token embedding module to get local tokens. Then, these tokens are fed to the global token embedding module which consists of three convolutional layers and one linear layer. Both local and global tokens are flattened into vectors of length of 96. In the following encoder, Global-Local Correlation Module (GLC), and decoder blocks, the number of local tokens is either downsampled or upsampled, while the number of global tokens remains as one. Hence we write the number of tokens in the output size as \emph{(1 global token + number of local tokens)}. After generating the output from decoder block4, a convolutional layer is applied only on the local tokens to compress the 8 channels to 1. We then convert this to a probability distribution by applying softmax to each frame.


\section{Experiments on Action Recognition}
\label{sec:action}

In addition to egocentric gaze estimation in the main paper, we also examine the application of our GLC module to the egocentric video action recognition task, and find that our method performs competitively with methods designed specifically for this task on EGTEA Gaze+. To this end, we remove the decoder in the gaze estimation model and keep only the visual token embedding, transformer encoder, and GLC modules. Generally, there are two ways to obtain activity class category prediction: adding a class embedding token at the first layer of transformer, or using pooling across all global tokens to obtain a final embedding. Then a fully-connected layer followed by softmax is used to predict probabilities for each category. We implement both strategies and compare our approaches with previous works in Table~\ref{tab:action_egtea}. We conduct these experiments only on EGTEA Gaze+~\cite{li2018eye} using the same split as gaze estimation. Note that the Ego4D~\cite{grauman2022ego4d} social benchmark does not contain action labels.

\begin{table}[tbp]
\tablestyle{8pt}{1.0}
\begin{center}
\begin{tabular}{lccccc}

\toprule
Methods & Cls Token & Pooling & Top1-Acc & Top5-Acc & Mean Cls Acc \\
\midrule
MViT~\cite{fan2021multiscale}          & \checkmark &            & 64.64 & 89.22 & 54.02 \\
MViT~\cite{fan2021multiscale}          &            & \checkmark & 63.45 & 88.72 & 55.34\\
MViT + Global Token                    & \checkmark &            & 64.44 & 88.72 & 55.28 \\
MViT + Global Token                    &            & \checkmark & 63.06 & 88.53 & 54.15 \\
\rowcolor[HTML]{FAEBD7} MViT + Global Token + GLC & \checkmark &            & 64.79 & 88.67 & 56.77 \\
\rowcolor[HTML]{FAEBD7} MViT + Global Token + GLC &            & \checkmark & 65.33 & \textbf{89.12} & \textbf{57.26} \\
\bottomrule
\end{tabular}
\end{center}
\caption{Results of action recognition on EGTEA Gaze+. We implemented two methods for classification -- adding an additional class token or using global average pooling. ``-"" means the result is unavailable. The complete models are highlighted.}
\label{tab:action_egtea}
\end{table}

For vanilla MViT~\cite{fan2021multiscale}, class token embedding performs better than the pooling operation. For both methods, simply adding global embedding has a minor influence on the overall performance ($-0.2\%$ on top1 accuracy, $-0.5\%$ on top5 accuracy and $+1.32\%$ on mean class accuracy while using the class token, and $-0.39\%$, on top1 accuracy, $-0.19\%$ on top5 accuracy and $-1.19\%$ on mean class accuracy while using pooling layer). This result suggests that simply adding global context as an additional token has minor influence on the action recognition performance.

In addition, adding our GLC module can only improve the model performance by a small margin when using class token embedding to predict action classes. We hypothesize that this is because only the class token is input into the linear layer for final prediction and re-weighted tokens from GLC are left unused. In contrast, when applying global average pooling on all local tokens, GLC improves top1, top5 and mean class accuracy over the counterpart that doesn't use GLC (\emph{MViT+Global Token}) by $+2.27\%$, $+0.59\%$ and $+3.11\%$, respectively. Gains over corresponding the MViT baseline are $+1.88\%$, $+0.4\%$ and $+1.92\%$ on the three metrics. These results indicate our proposed GLC module is a robust and general design that also improves the action recognition performance. However, the impact on action recognition is smaller compared with egocentric gaze estimation. 

We note that our model achieves a competitive performance for action recognition on EGTEA Gaze+ without additional design for this specific task. Our top1 accuracy of $65.33\%$ exceeds Wang et al. (2020)~\cite{wang2020symbiotic} by $+1.23\%$, and is only a $-1.17\%$ difference from Hao et al. (2022)'s~\cite{hao2022group} recent state-of-the-art method for this benchmark of $66.5\%$. We also want to emphasize that we conduct these action recognition experiments to demonstrate the generalization ability of our proposed GLC module rather than aim to produce SOTA results on action recognition.


\begin{figure}[t]
\centering
\includegraphics[width=\linewidth]{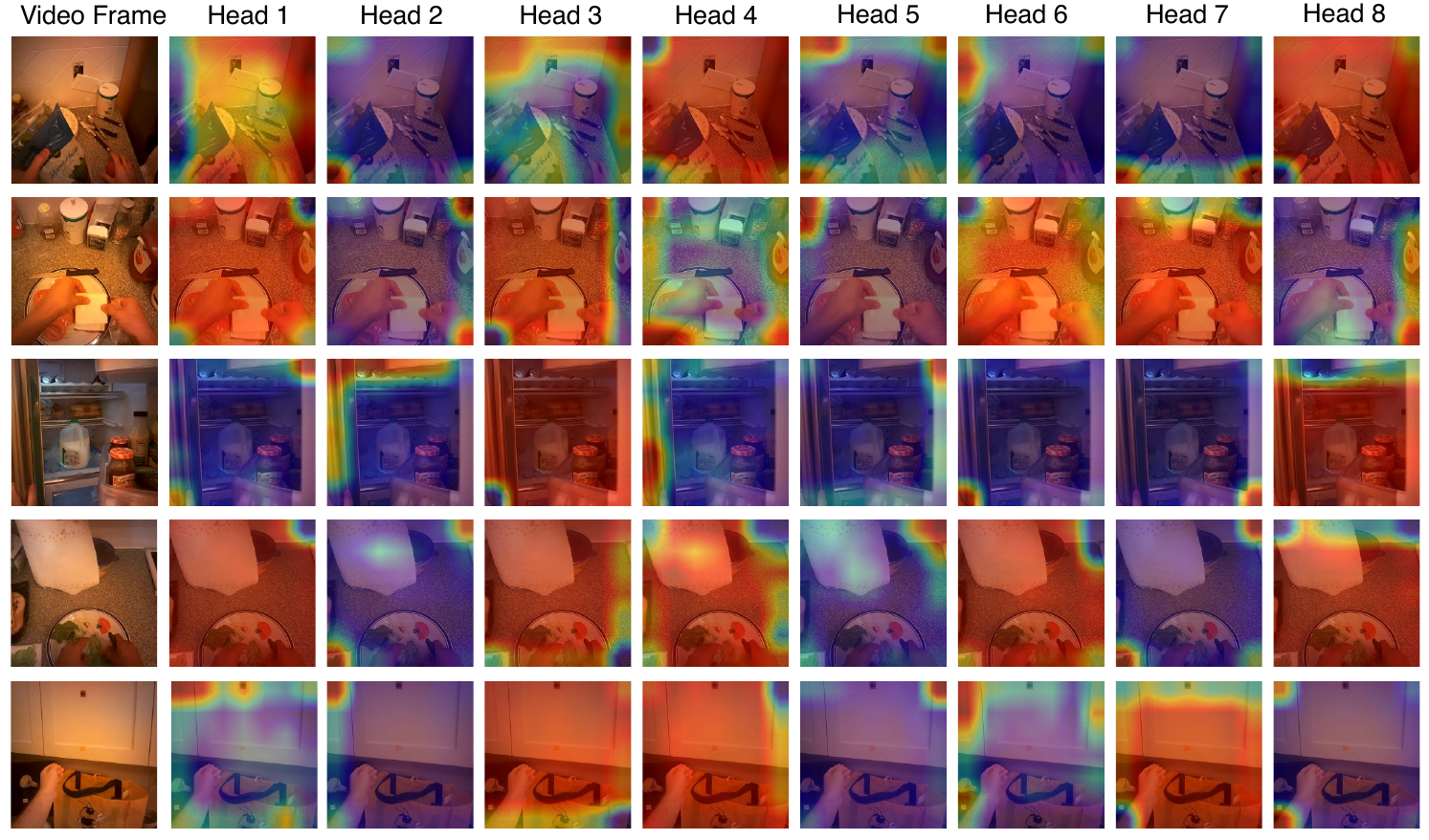}
\caption{Visualization of the eight heads in global-local correlation module for action recognition.}
\label{fig:action_glc_vis}
\end{figure}

Additionally, we visualize the global-local correlation weights of the GLC in Fig.~\ref{fig:action_glc_vis}. Importantly, the learned global-local correlation is vastly different from the gaze distribution when the model is trained for action recognition; in contrast, a stronger connection between the learned global-local correlation and gaze distribution can be observed when the model is trained for gaze estimation (see Fig.~\ref{fig:suppl_glc_vis}). How to design a weakly-supervised model for egocentric gaze estimation remains an open question.



\section{Details of Different Global Visual Embedding Strategies}
\label{sec:embed}
\begin{figure}[t]
\centering
\includegraphics[width=0.9\linewidth]{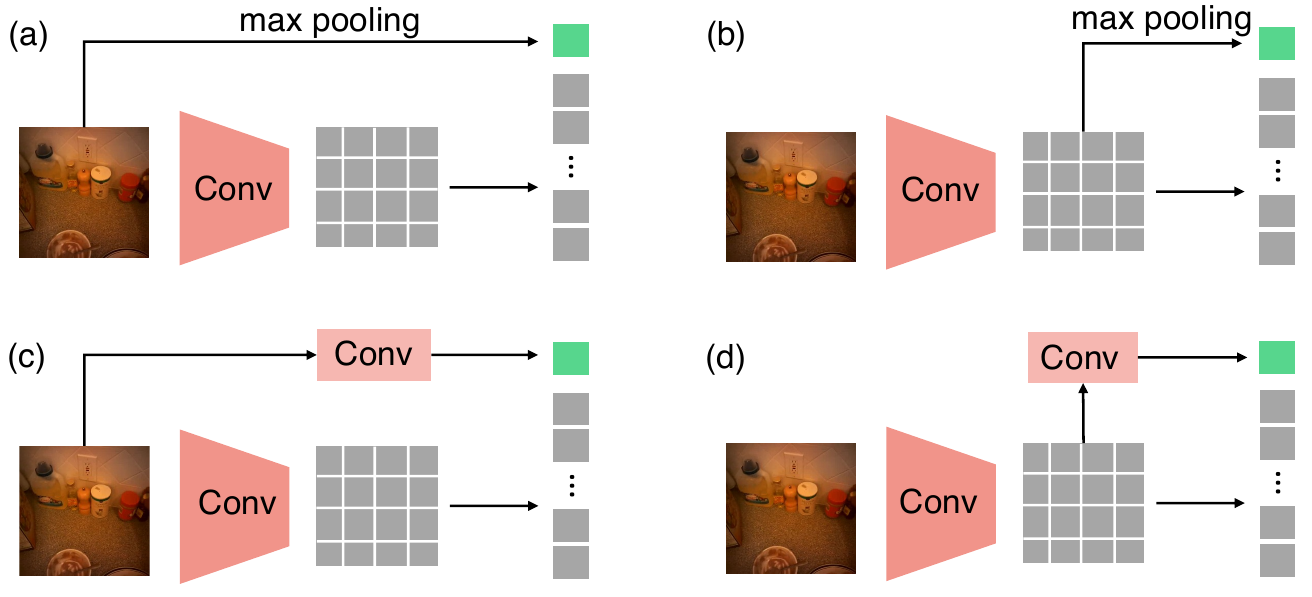}
\caption{Four different approaches of global visual token embedding.}
\label{fig:embed}
\end{figure}
We present further details of the four global visual embedding strategies we studied in Section 4.2 of the main paper. As demonstrated in Fig.~\ref{fig:embed}, (a) implements max pooling on input frames directly, and (b) implements max pooling on local visual tokens. For (c) and (d), we replace max pooling operations in (a) and (b) with a sequence of convolutional layers. The specific parameters of (d) are detailed in Table~\ref{tab:architecture}. For global embedding in (c), input video frames are fed into a convolutional layer that is identical to the layer used for local token embedding (\emph{i.e.}, kernel is $3\times7\times7$ and stride is $2\times4\times4$.) Then, the output is passed to a sequence of convolutional layers identical to (d).

\section{More Visualization Examples of Gaze Estimation}
\label{sec:visgaze}

\begin{figure}[tbp]
\centering
\includegraphics[width=0.9\linewidth]{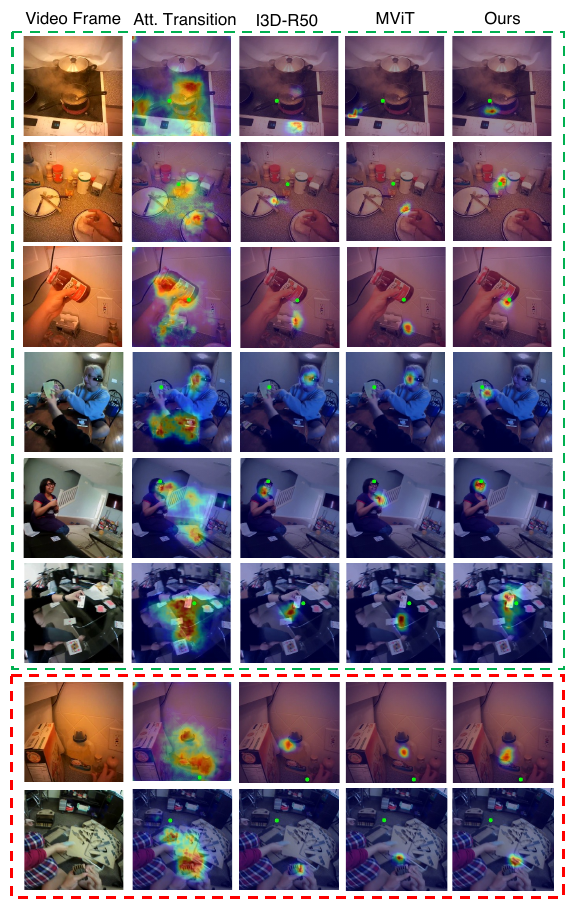}
\caption{Visualization of gaze estimation. Both successful cases (in green box) and failure cases (in red box) of our model are demonstrated. Green dots present ground truth.}
\label{fig:suppl_cmp_sota}
\end{figure}

More visualizations of gaze prediction of both our model and previous state-of-the-art approaches are presented in Fig.~\ref{fig:suppl_cmp_sota}. Our proposed model can accurately predict the gaze distribution even when the scene context is very complicated, while the other three approaches may be misled by background objects or produce predictions with too much uncertainty.

\begin{figure}[tbp]
\centering
\includegraphics[width=\linewidth]{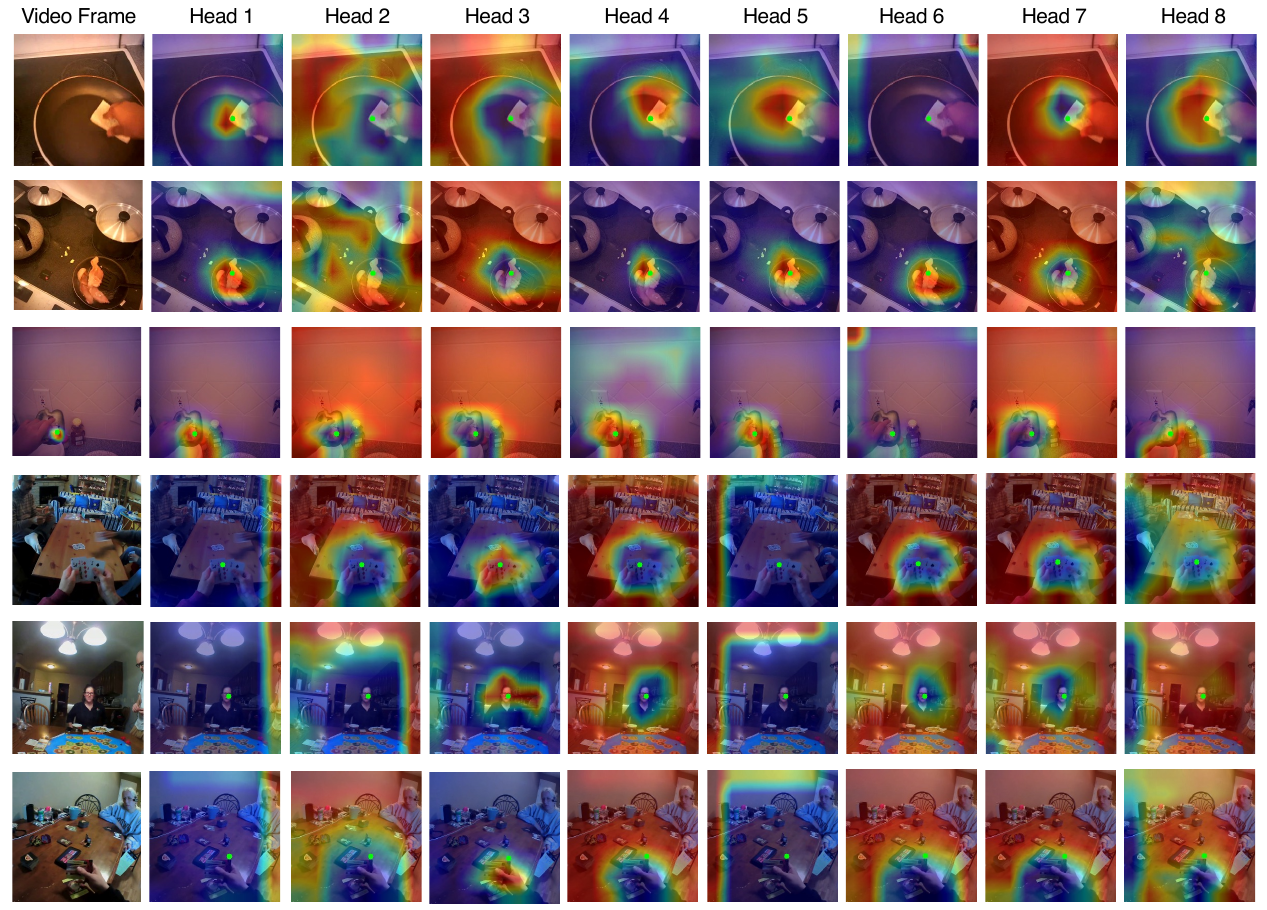}
\caption{Visualization of the eight heads in the Global-Local Correlation module. Green dots represent the ground truth.}
\label{fig:suppl_glc_vis}
\end{figure}

We provide more examples of GLC visualizations in Fig.~\ref{fig:suppl_glc_vis}. The 8 heads capture features of different areas which is consistent with the examples in the main paper. On the EGTEA Gaze+ dataset, the maps produced by heads 1, 4, 5, and 8 highlight pixels around the gaze point with different uncertainty (which is illustrated by the size of highlighted area). The other four heads focus on surrounding objects and leave gaze areas unattended. As for the Ego4D data, only head 3 captures the wearers' attention, while the other heads fully focus on the backgrounds in different aspects. This supports our key conclusion in the main paper that our GLC module learns to model human attention by setting different weights from local to global tokens, capturing many facets of scene information (both around the gaze target and in the background) in the multi-headed attention mechanism.

\section{Future Work}
\label{sec:future}
In this paper, we studied the explicit integration of global scene context for egocentric gaze estimation and proposed a novel modeling approach for this problem. We also showed the results of our proposed architecture on egocentric action recognition in this supplementary material to demonstrate our model's generalization ability. Our findings also point to several exciting future research directions:

\begin{itemize}
    \item Our proposed GLC module has the potential to address other video understanding tasks including visual saliency prediction in third-person video, active object detection, and future forecasting. We plan to study the effect of our method on those tasks in our future work.
    
    \item Our modeling work can be expanded to understanding human gaze behavior associated with multiple sensing modalities, especially in the social conversation setting. An exciting future direction is incorporating audio signals into egocentric gaze estimation.
    
    \item  Our proposed GLC fails to learn the gaze distribution when the model is trained to predict the action labels. How to design a weakly supervised model for egocentric gaze estimation using action labels is an interesting problem.
\end{itemize}

\begin{table}[t]
\begin{center}
\renewcommand\arraystretch{1.2}
\begin{tabular}{c|c|c}
\hline
Stages  & Operators & Output Size \\
\hline
data            & -  & $8\times256\times256$ \\
\hline
local token embedding & $\begin{array}{c} Conv(3\times7\times7,\ 96) \\ stride\ 2\times4\times4 \end{array}$ & $96\times4\times64\times64$ \\
\hline
global token embedding & $\begin{array}{c} Conv(3\times3\times3,\ 96) \\ Conv(3\times3\times3,\ 96) \\ Conv(3\times3\times3,\ 96) \\ Linear(24576) \\ stride\ of\ each\ conv\ 1\times2\times2 \end{array}$ & $96\times1$ \\
\hline
tokenization & flattening and concatenation & $96\times (1+4\times64\times64)$ \\
\hline
encoder block1          & $\left[\begin{array}{c} MSA(96) \\ MLP(384)\end{array}\right]\times1$ & $192\times(1+4\times64\times64)$ \\
\hline
encoder block2          & $\left[\begin{array}{c} MSA(192) \\ MLP(768)\end{array}\right]\times2$ & $384\times(1+4\times32\times32)$ \\
\hline
encoder block3          & $\left[\begin{array}{c} MSA(384) \\ MLP(1536)\end{array}\right]\times11$ & $768\times(1+4\times16\times16)$ \\
\hline
encoder block4          & $\left[\begin{array}{c} MSA(768) \\ MLP(3072)\end{array}\right]\times2$ & $768\times(1+4\times8\times8)$ \\
\hline
global-local correlation & $\left[\begin{array}{c} GLC(768) \\ MLP(3072) \\ concatenation\ in\ channel\end{array}\right]\times1$ & $1536\times(1+4\times8\times8)$ \\
\hline
decoder block1          & $\left[\begin{array}{c} MSA(1536) \\ MLP(3072)\end{array}\right]\times1$ & $768\times(1+4\times16\times16)$ \\
\hline
decoder block2          & $\left[\begin{array}{c} MSA(768) \\ MLP(1536)\end{array}\right]\times1$ & $384\times(1+4\times32\times32)$ \\
\hline
decoder block3          & $\left[\begin{array}{c} MSA(384) \\ MLP(768)\end{array}\right]\times1$ & $192\times(1+4\times64\times64)$ \\
\hline
decoder block4          & $\left[\begin{array}{c} MSA(192) \\ MLP(384)\end{array}\right]\times1$ & $96\times(1+8\times64\times64)$ \\
\hline
head                    & $\begin{array}{c} Conv(1\times1\times1,\ 1) \\ stride\ 1\times1\times1 \end{array}$ & $8\times64\times64$ \\
\hline
\end{tabular}
\end{center}
\caption{Architecture of the proposed model. Convolutional layers are denoted as $Conv(kernel\ size,\ output\ channels)$. Numbers of input channels of multi-head self-attention are shown in the parenthesis of $MSA$. Dimensions of the hidden layer in multi-layer perceptrons are listed in parenthesis of $MLP$. In tokenization, local and global tokens are reshaped and concatenated. In global-local correlation, the output is concatenated with its input in the channel dimension. Head only takes local tokens as input.}
\label{tab:architecture}
\end{table}

\end{document}